\documentclass[10pt,twocolumn,letterpaper]{article}

\usepackage{iccv}
\usepackage{times}
\usepackage{epsfig}
\usepackage{graphicx}
\usepackage{amsmath}
\usepackage{amssymb}
\usepackage{algorithm}
\usepackage{algorithmic}

\usepackage{subcaption}
\DeclareMathOperator*{\argmax}{argmax} 
\usepackage[pagebackref=true,breaklinks=true,letterpaper=true,colorlinks,bookmarks=false]{hyperref}

\iccvfinalcopy

\ificcvfinal\fi
\begin{document}

\title{Learning from Adversarial Features for Few-Shot Classification}

\author{Wei Shen \qquad Ziqiang Shi \qquad Jun Sun \\
Fujitsu Research \& Development Center, Beijing, China.
\\
{\tt\small \{shenwei, shiziqiang, sunjun\}@cn.fujitsu.com}
}

\maketitle

\begin{abstract}
   Many recent few-shot learning methods concentrate on designing novel model architectures. In this paper, we instead show that with a simple backbone convolutional network we can even surpass state-of-the-art classification accuracy. The essential part that contributes to this superior performance is an adversarial feature learning strategy that improves the generalization capability of our model. 
   In this work, adversarial features are those features that can cause the classifier  uncertain about its prediction.
   In order to generate adversarial features, we firstly locate adversarial regions based on the derivative of the entropy with respect to an averaging mask. Then we use the adversarial region attention to aggregate the feature maps to obtain the adversarial features. 
   In this way, we can explore and exploit the entire spatial area of the feature maps to mine more diverse discriminative knowledge. 
   We perform extensive model evaluations and analyses on miniImageNet and tieredImageNet datasets demonstrating the effectiveness of the proposed method. 
   
\end{abstract}

\section{Introduction}
Few-shot classification aims at classifying \textit{query} images into classes each of which has a few labelled \textit{support} images. The number of classes is called \textit{way} and the number of support images is called \textit{shot}. For example, if a task contains 5 classes and each class has one labelled image, it is a 5-way 1-shot task. If each class has five labelled images, it is a 5-way 5-shot task. The main challenge of few-shot classification is that the classes in the test set have no overlap with those in the training set. Thus, it requires models to be able to generalize to novel classes rather than simply over-fit the training data. 

Recent years have witnessed rapid advances in few-shot learning. Meta-learning based methods address the challenge by learning meta knowledge from a number of different tasks and adapting the knowledge to novel tasks~\cite{vinyals2016matching,finn2017model,snell2017prototypical}. The tasks in the training phase usually mimic the settings that will be used in the test phase to minimize the gap between training settings and test settings. The assumption is that the knowledge shared among those training tasks can be transfered to other novel tasks. Those shared knowledge can be a metric space~\cite{snell2017prototypical}, a good initialization~\cite{finn2017model} or task representations~\cite{oreshkin2018tadam} \etc. 
Although those methods have achieved improved results, there are two counter intuitive points which need to be noticed. (a) As pointed out in ~\cite{snell2017prototypical}, if the test tasks are 5-shot tasks but the model is trained on 1-shot tasks the performance will be much lower than the model trained on 5-shot tasks and vice versa. It means that we have to train one model for 1-shot and another for 5-shot tasks in order to achieve good performance on both tasks. 
(b) If the way of the tasks we use to train the model is the same as those of the tasks in the test phase, we will not achieve the highest classification accuracy either. Instead we have to increase the number of ways (\eg to 20-way) of training tasks to get improved classification accuracy on test tasks (\eg 5-way). 

Another line of current work is to dynamically generate classifier's weights for novel classes. Given a pretrained model, one can compare the feature of a novel class to those of the known classes and inference the weight for the novel class using attention mechanism~\cite{gidaris2018dynamic}. However, the training phase has to be usually split into two stages, one for feature extraction and the other for weight generation. One can also train a model to learn to generate different features from the distribution of a given class on the training data and perform such augmentation for novel classes with few samples~\cite{schwartz2018delta}. Thus, in the test phase the classifier can be trained with more data. Nevertheless, generating those samples and training a new classifier is time consuming. 

Although the above mentioned methods have achieve impressive results, few of them pay much attention to the quality of the basic feature extractor. Training neural networks in an end-to-end way sometimes results in models over-fitting the training data. 
For example, if a classifier is trained on a dog dataset, the classifier may pay much attention to the head of the dog and less to the body even if the body may also contain discriminative information~\cite{zhang2018adversarial}.  Another extreme example is described as blind spots in~\cite{lakkaraju2017identifying} where a predictive model trained on images of black dogs and white cats will incorrectly label a white dog (test image) as a cat with high confidence. The blind spot example indicates that the model may merely learn some conspicuous discriminative features and does not dive deep into the intrinsic characteristic of the target object. This problem is more obvious in the scenario of few-shot learning. Since test classes are different from training classes, if the model over-fits the training data, its performance on novel classes will be poor.

\begin{figure*}
\centering
\includegraphics[width=1\linewidth]{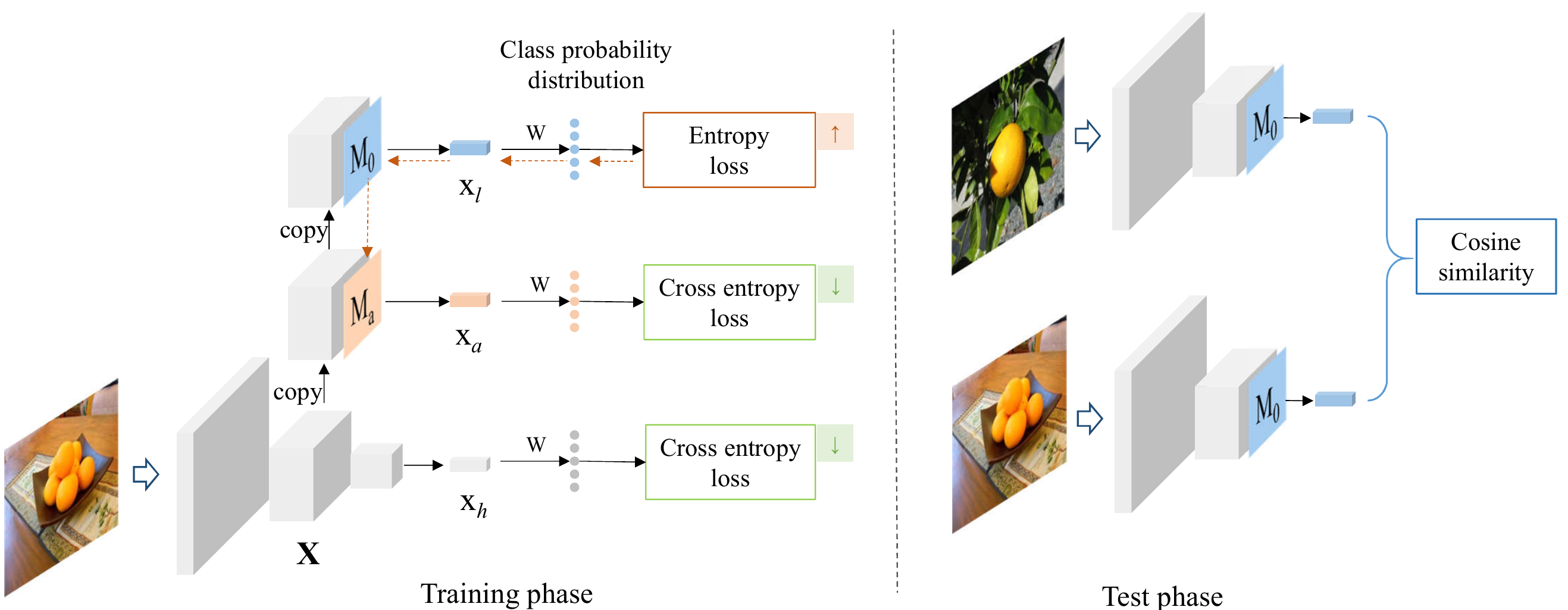}
\caption{The pipeline of the proposed method in the training phase (left) and the test phase (right). In the training phase, we extract three feature vectors from low level feature maps $\mathbf{X}$, \ie the global average pooling feature $\mathbf{x}_l$ (1st row), the adversarial feature $\mathbf{x}_a$ (2nd row) and the high level feature $\mathbf{x}_h$ (3rd row). $\mathbf{W}$ is the weight matrix of the classifier. The gradient of the entropy with respect to $\mathbf{M}_0$ is back propagate to $\mathbf{M}_0$ and then is used to obtain the adversarial region attention (orange line). The two cross entropy losses are used to train the model and update the parameters of the entire network. In the test phase, we only extract the global average pooling features from the low level feature maps as representations of images. Cosine similarity is used as the metric to measure the similarity between two images. (Best viewed in color.)
}
\label{fig:pipeline}
\end{figure*}

In this work, we show that models trained on adversarially perturbed features can generalize better compared to those trained on clean features. Even without further adaptation on novel tasks, we can still surpass current sate-of-the-art classification accuracy on few-shot classification tasks. 
The pipeline is shown in Figure~\ref{fig:pipeline}. In the training phase, we feed images to our model and obtain the convolutional feature maps. Based on the feature maps, we extract three feature representations, \ie the global pooling feature, the adversarial feature and the high level feature. The last two features are used to construct multi-scale classification loss that is used to update the parameters of our model while the first one is used to generate adversarial attention mask. By incorporating the adversarial mask, we force the classifier to leave its comfortable zone and focus on other regions. In this way, our model can explore and exploit the overall feature maps in-depth to learn more diverse discriminative information compared to traditional training scheme. 
In the test phase, only the global average pooling features are used for object representation.

The contributions of our work are as follows.
\begin{itemize}
\item We propose a feature learning scheme based on adversarial features. Our model spatially explores and exploits the feature maps to mine discriminative information. The learned feature extractor can be directly used for few-shot classification without any adaptation on novel tasks.

\item We propose to use a multi-scale classifier to generate the adversarial attention. We show that compared to the classifier trained on single scale features, the attention region of multi-scale classifier is more accurately focused on the object.

\item We show that with the proposed method we can use the same trained model for both 5-way 1-shot tasks and 5-way 5-shot tasks and achieve new state-of-the art results on both tasks. It demonstrates that our model indeed learns an efficient metric space that generalize well on novel tasks.
\end{itemize}

\section{Related work}
\subsection{Few-shot learning}
In this section, we roughly categorize recent few-shot learning methods into two categories, \ie meta-learning based approaches and weight generation based approaches.

\noindent \textbf{Meta-learning based approaches}.
Models in this group are typically trained on episodes of tasks~\cite{ravi2016optimization,rusu2018metalearning,sung2018learning,vinyals2016matching}. Snell~\etal introduced prototypical networks that learned a metric space in which classification could be performed by computing distances to prototype representations of each class~\cite{snell2017prototypical}. Sung \etal proposed relation networks that learned a deep distance metric to compare images within episodes~\cite{sung2018learning}. Boris~\etal proposed to learn a task-dependent metric space where the task representation was the mean of the class prototypes~\cite{oreshkin2018tadam}.

\noindent \textbf{Weight generation based approaches}.
Models in this group learn to generate classifier's weights for novel classes~\cite{gidaris2018dynamic,qi2018low,qiao2018few,schwartz2018delta,wang2018low,zhang2018metagan}.
Spyros~\etal devised a model that was able to efficiently learn novel categories
from only a few training data while not forgetting the initial classes on which it was trained~\cite{gidaris2018dynamic}. Huang~\etal directly set weights for new classes based on an appropriately scaled copy of the embedding layer activations~\cite{qi2018low}. Siyuan~\etal also analyzed the relationship between the parameters and the activations and  proposed to adapt a pre-trained network to novel categories by predicting the parameters from the activations~\cite{qiao2018few}. Instead of directly predicting classifier's weight, Schwartz~\etal proposed to learn to extract transferable intra-class deformations between same-class pairs and to apply those deformations to examples from a novel class~\cite{schwartz2018delta}. Ruixiang~\etal augmented
vanilla few-shot classification models with the ability to discriminate between real and fake data so that the decision boundary became much sharper leading to better generalization~\cite{zhang2018metagan}. 

\subsection{Adversarial learning}
Adversarial learning is becoming more and more popular in recent years. In this section, we briefly discuss the most relevant work~\ie adversarial attack and adversarial complementary learning.

\noindent \textbf{Adversarial attack}.  Adversarial examples are referred to inputs formed by applying small but intentionally worst-case perturbations resulting in the model outputting an incorrect answer~\cite{goodfellow2014explaining}. Adversarial example generation approaches can find the flaw of a machine learning model and then attack the model. 
There are many ways to generate adversarial perturbations~\cite{papernot2017practical,moosavi2016deepfool,papernot2016limitations,goodfellow2014explaining,szegedy2013intriguing,he2018decision}. Ian~\etal designed a gradient sign method (FGSM) to generate the perturbation according to the sign of the gradient of the cost function with respect to the input~\cite{goodfellow2014explaining}. Nicolas~\etal proposed Jacobian-based Saliency Map Attack (JSMA) to construct adversarial saliency maps enabling an efficient exploration of the adversarial-samples search space~\cite{papernot2016limitations}.

\noindent \textbf{Adversarial complementary learning}
is introduced in ~\cite{wei2017object,zhang2018adversarial} for object localization. In adversarial complementary
learning, two or more classifiers are trained to progressively mine discriminative object regions. Yunchao~\etal proposed to use three classification networks to sequentially discover complement object regions by erasing the current mined regions~\cite{wei2017object}. Xiaolin~\etal integrated three networks in~\cite{wei2017object} into a single network and generated the localization map by forwarding the network only once.

\section{Method}
In this section, we will give a detailed description of our method. Firstly, we will show how to obtain adversarial features from low level feature maps. Then we will introduce multi-scale feature learning. Finally, we will provide some implementation details.

\subsection{Adversarial features generation}
\subsubsection{Adversarial goal}
Given feature maps $\mathbf{X}$ with size $C\times H \times W$ and a classifier $f$, we are trying to find an attention mask $\mathbf{M}_a$ with which the linear combination of feature vectors in $\mathbf{X}$ can cause the classifier uncertain about its predictions. We refer to $\mathbf{M}_a$ as the adversarial region attention. Formally, 
\begin{equation}
\mathbf{M}_a=\argmax_{M}(e(f(\mathbf{X}, \mathbf{M}), y)),
\label{eq:ma_y}
\end{equation}
or 
\begin{equation}
\mathbf{M}_a=\argmax_{M}(e(f(\mathbf{X}, \mathbf{M}))),
\label{eq:ma_noy}
\end{equation}
depending on whether the ground-truth label $y$ of the input image is used. $e(.)$ is the metric that measures the uncertainty of the classifier's prediction. In this work, we will use entropy as default to measure the uncertainty of the prediction. The entropy can be written as  
\begin{equation}
e(f(\mathbf{X}, \mathbf{M}_a))=\Sigma_{i=1}^{n}-p_ilog(p_i),
\label{eq:ms}
\end{equation}
where $p_i$ is the prediction probability of the $i$-th class and $n$ is the number of classes.

\subsubsection{Adversarial features}
\label{sec:adv_feat}
One can strictly adhere to  Equation~\eqref{eq:ma_y} or ~\eqref{eq:ma_noy} to calculate the analytic solution of $\mathbf{M}_a$. However, we empirically observe that one single gradient step applied on an averaging mask works surprisingly well in all our scenarios. 
In order to obtain $\mathbf{M}_a$ using back propagation, we firstly describe the forward pass. To reduce the dimension of convolution feature maps, we apply global average pooling on feature maps as in~\cite{he2016deep}. Global average pooling spatially averages layer activations and outputs a feature vector. It can also be explicitly implemented by applying an averaging mask $\mathbf{M}_0$ with each element be $1/(H\times W)$ on the feature maps and sum up the activations as 
\begin{equation}
\mathbf{x}_l=\Sigma_{i,j}\mathbf{X}\mathbf{M}_0,
\label{eq:gap}
\end{equation}
where $(i,j)$ indicates the spatial location on the feature maps. Once we have the global average pooling feature vector $\mathbf{x}_l$, we feed it to the classifier $f$ which will output a probability distribution over the training classes. If the classifier is very confident about its prediction, the distribution will have a single peak and otherwise the distribution will be flat. 
We try to find feature vectors from the feature maps that can flatten the probability distribution. Since we use entropy $l_{ent}$ to indicate uncertainty, we back propagate the gradient of $l_{ent}$ with respect to the mask $\mathbf{M}_0$. Then we can obtain the adversarial region attention mask $\mathbf{M}_a$ by updating $\mathbf{M}_0$ as

\begin{equation}
\mathbf{M}_a=\mathbf{M}_0+\gamma \triangle \mathbf{M},
\label{eq:ma_g}
\end{equation}
\begin{equation}
\triangle \mathbf{M}=\frac{\partial l_{ent}}{\partial \mathbf{M}}|_{\mathbf{M}_0}
\label{eq:tr_m}
\end{equation}
where $\gamma$ is the step size. 
Given $\mathbf{M}_a$, we can directly calculate the adversarial feature $x_a$ as
\begin{equation}
\begin{split}
\mathbf{x}_a&=\Sigma_{i,j}\mathbf{M}_a\mathbf{X} \\
&=\Sigma_{i,j}(\mathbf{M}_0+\gamma \triangle \mathbf{M})\mathbf{X} \\
&=\Sigma_{i,j}\mathbf{M}_0\mathbf{X}+\gamma \Sigma_{i,j}\triangle \mathbf{M}\mathbf{X} \\
&=\mathbf{x}_l+\gamma \triangle \mathbf{x}_l.
\end{split}
\label{eq:adv_noise}
\end{equation}
From Equation~\eqref{eq:adv_noise}, we can find that adversarial feature is a combination of an averaged feature representation and a small perturbation of adversarial noise. Therefore, models trained on those perturbed features will have lower risk of over-fitting the training data and the learned metric space will also be smooth which is helpful for knowledge transfer.

\subsubsection{An intuitive understanding}
Suppose we have a classifier trained on the average pooling feature vectors $\mathbf{x}_l$. The classifier can be sensitive to features in some conspicuous discriminative regions. If the feature activations in those regions dominate the average representation, the classifier will be very confident about its prediction and it will enter its comfort zone and stop learning other discriminative information. Mathematically speaking, in this case we are in the saturation area of the log softmax where the gradient magnitude is very small. Fortunately, we are able to find adversarial regions on the feature maps that can reduce our model's confidence. Then we can force our model to learn from those regions and thus the discriminative information learned by our model is more diverse than that learned by conventional models. With diverse discriminative knowledge, our model can generalize well on novel tasks.

\subsubsection{Design choice}
\noindent \textbf{Adversarial features from feature maps}.
In adversarial attack, researchers usually try to find subtle perturbations in image space that cause the classifier to make incorrect predictions while our adversarial features are calculated based on convolutional feature maps. The reason is that we are focusing on forcing our model to learn diverse semantic discriminative information which is helpful for generalization. However, subtle changes in image space does not provide any semantically different information. On the contrary, our adversarial region attention can be applied on convolutional feature maps to dynamically generate semantically different adversarial features according to Equation~\ref{eq:adv_noise}.

\noindent\textbf{Entropy loss}
The reason to choose entropy loss instead of cross entropy loss in adversarial attention generation is that it offers much more abundant information in back propagation. For cross entropy loss, when the classifier is well trained, the gradient magnitude with respect to the mask will be too small for efficient learning.

\subsection{Multi-scale feature learning}
\subsubsection{High level feature learning} 
Convolutional networks extract high level structured information at high layers~\cite{zeiler2014visualizing}. We train a fully convoutional neural network whose output feature maps size is $C \times 1 \times 1$ where $C$ is the number of channels. We denote this feature vector as $\mathbf{x}_h$. It contains the overall image information since it gradually extracts more and more hierarchical information from the entire image. 
The classification loss for this level is the cross entropy loss 
\begin{equation}
\begin{split}
l_h &= -\text{log}p_y(\mathbf{x}_h), 
\end{split}
\end{equation}
where $y$ is the label of the input sample and $p_y$ is the softmax probability. 

\subsubsection{Low level feature learning} 
In few shot learning, the classes in the test phase and those in the training phase are disjoint. Since high level features more concentrate on class-specific knowledge, the activations from the last layer cannot be directly used as the representations of novel categories. 
A common practice is to use the activations from intermediate layers to avoid the dataset bias. The assumption is that even though the categories are disjoint between the training data and the test data, they may still share some common local patterns. The activations from the intermediate layers are sensitive to those local patterns and thus are potentially transferable for novel classes.

Given the feature maps $\mathbf{X}$ from the intermediate layer. One can train a classifier based on the aggregated feature from $\mathbf{X}$.
In this work, we use the adversarial attention mask $\mathbf{M}_a$ for aggregation. The training loss is also the cross entropy loss

\begin{equation}
l_l = -\text{log}p_y(\mathbf{x}_a).
\label{eq:loss_low}
\end{equation}
Note that the adversarial mask is only used in the training phase for mining adversarial features. In the test phase, we apply global average pooling to obtain the feature vector.

\subsubsection{Multi-scale learning}
During training, the loss function is the combination of high level classification loss and low level adversarial classification loss
\begin{equation}
L=l_h+l_l.
\end{equation}
The entropy loss is not used for parameter update. It is only used to calculate the adversarial mask $\mathbf{M}_a$. 

Despite the generalization capability of low level features, the relatively small receptive field of low level activations may cause the model incidentally capture non-object regions as the representative features of the class. This happens especially when global pooling is applied to the last  convolutional feature maps. The reason is that global pooling removes spatial information from the feature maps and only local activations are aggregated for classification.
In contrast, the receptive field of high level activations covers the entire image. Therefore, it is reasonable to assume that the features learned at high level to be more accurate.
To improve discriminative feature learning in low level layers, we share the classifier between high level feature learning and low level feature learning. This simple multi-scale learning strategy aligns features from different scales by enabling the knowledge flow between high level representations and the low level representations. The benefit of the multi-scale learning strategy is shown in Figure~\ref{fig:grad_57}.

\subsubsection{Design choice}
\noindent\textbf{Cosine similarity}.
For both the entropy loss and the cross entropy loss, we use the cosine similarity between the feature representations and the classifier's weight vectors instead of the dot product as the distance metric before softmax activations. Cosine similarity based models are demonstrated to generalize significantly better on novel categories~\cite{gidaris2018dynamic}. However, the range of cosine similarity is fixed to [-1, 1] which is difficult for efficient learning. A common practice it to multiply the similarity value with a scaling factor $s$ which can be a fix value~\cite{deng2018arcface} or a learnable parameter~\cite{wang2017normface} to control the peakiness of the probability distribution. In this work we fix the scaling factors due to its simplicity in implementation.

\noindent\textbf{Step size $\gamma$} in Equation~\ref{eq:ma_g} and~\ref{eq:adv_noise}
is just like the learning rate when training deep models. It cannot be too large or too small. Since it is multiplied to the gradient from the entropy loss, we assume there exists a reciprocal relationship between $\gamma$ and the scaling factor $s$ for stable training. In other words, if the scaling factor $s$ is large, we should choose a small $\gamma$ and otherwise we choose a large value. Therefore, we have
\begin{equation}
\gamma = 1/s.
\end{equation}
Note that we may not strictly follow this reciprocal relationship. A slightly different value does not effect the performance of our model much (see Section~\ref{sec:gamma} for details).

The pseudo code of the proposed method is provided in Algorithm~\ref{alg}.
\begin{algorithm}
  \caption{Learning from adversarial features}
  \label{alg}
  \begin{algorithmic}
  \REQUIRE Low level feature maps $\mathbf{X}$, a classifier $f$ and an averaging mask $\mathbf{M}_0$.
  \STATE $\mathbf{x}_l$ $\gets$ GlobalAveragePool($\mathbf{X}$, $\mathbf{M}_0$)
  \STATE $l_{ent}$ $\gets$ EntropyLoss($f(\mathbf{x}_l)$)
  \STATE $\triangle \mathbf{M} \gets$ Back propagate the partial derivative $\frac{\partial l_{ent}}{\partial \mathbf{M}}|_{\mathbf{M}_0}$
  \STATE $\mathbf{M}_a$ $\gets$ $\mathbf{M}_0+\gamma \triangle \mathbf{M}$
  \STATE $\mathbf{x}_a \gets \Sigma_{i,j}\mathbf{M}_a\mathbf{X}$
  \STATE $l_l \gets$ CrossEntropyLoss($f(\mathbf{x}_a)$)
  \STATE $\mathbf{x}_h \gets$ Forward $\mathbf{X}$ to another conv-pool block
  \STATE $l_h \gets$ CrossEntropyLoss($f(\mathbf{x}_h)$)
  \STATE Update network parameters to minimize $l_l+l_h$.
  \end{algorithmic}
\end{algorithm}

\subsection{Implementation details}
For both miniImageNet dataset and tieredImageNet dataset we use the same network and the same hyper-parameter settings. The learning rate is set to 0.001 as~\cite{snell2017prototypical} and is halved every 10 epochs. We train our model for 50 epochs and choose the model that achieves the best 5-way 1-shot classification accuracy on validation set for testing. Note that our model is trained just like common classification models whose input is a batch of images rather than a batch of tasks used in meta-learning. We train our model once and test it on both 5-way 1-shot and 5-way 5-shot tasks. For 1-shot tasks, we compare the cosine similarity of the query image features to each support image feature and assign the label of the nearest neighbor to the query image. For 5-shot tasks, we average the feature vectors of the five images that belong to the same class as the prototype representation of that class. Then we assign the label of the nearest prototype to the query image.
The CNN backbone we use is a VGG-like~\cite{simonyan2014very} network. Details are shown in Table~\ref{tb:net}. We used activations from conv5 layer as the low level feature maps $\mathbf{X}$. The input image size is $128\times 128$ so that the spatial size of feature maps from the conv5 layer is $8\times 8$ which is a sufficient size for adversarial feature mining. The scaling factor for training cross entropy loss is fixed to 20 and that for adversarial region attention is fixed to 5 and thus $\gamma=1/s=0.2$. In the training phase, we only employ random flip as data augmentation. In the test phase, we randomly sample 1000 tasks and report the average accuracy. 

\section{Experiments}
In this section, we evaluate our method on both miniImageNet dataset and tieredImageNet dataset. We will compare our results with current state-of-the-art results on both 5-way 1-shot tasks and 5-way 5-shot tasks. We also perform ablation studies to show how much improvement is brought by adversarial feature learning and how much is brought by multi-scale feature learning. We also evaluate the vulnerability of different baseline models. 

\subsection{Datasets}
\textbf{miniImageNet}~\cite{vinyals2016matching} is a subset of ILSVRC-12~\cite{russakovsky2015imagenet}. It contains 100 classes with 600 images per class. We follow the split in ~\cite{ravi2016optimization}. There are 64, 16, 20 classes for training, validation and test.

\textbf{tieredImageNet}~\cite{ren2018meta} is a much larger subset of ILSVRC-12~\cite{russakovsky2015imagenet}. It contains 608 classes belonging to 34 categories grouped according to ImageNet~\cite{deng2009imagenet}. These categories are split into 20 training (351 classes), 6 validation (97 classes)  and 8 test (160 classes) categories. Unlike miniImageNet dataset, all the training classes in tieredImageNet dataset are sufficiently distinct from the test classes.

\begin{table}
\begin{center}
\begin{tabular}{c|c}
\hline
Layer & Network details\\
\hline\hline
conv1 & 2$\times$ Conv(128,3,3)-BN-leakyReLU \\
- & Maxpool(2,2) \\
conv2 & 2$\times$ Conv(128,3,3)-BN-leakyReLU \\
- & Maxpool(2,2) \\
conv3 & 2$\times$ Conv(256,3,3)-BN-leakyReLU \\
- & Maxpool(2,2) \\
conv4 & 2$\times$ Conv(512,3,3)-BN-leakyReLU \\
- & Maxpool(2,2) \\
conv5 & 2$\times$ Conv(512,3,3)-BN-leakyReLU \\
\hline
- & Maxpool(2,2) \\
conv6 & 2$\times$ Conv(512,3,3)-BN-leakyReLU  \\
- & Maxpool(2,2) \\
conv7 & Conv(512,2,2)-BN-leakyReLU \\
$f$ & FC \\
\hline
\end{tabular}
\end{center}
\caption{Network details. \lq\lq Conv($c$,3,3)" indicates there are $c$ convolution kernels with size $3\times 3$. 2$\times$ means there are two consecutive blocks.  \lq\lq Maxpool(2,2) " means max-pooling with stride 2 and pooling window size is $2\times 2$. \lq\lq BN " indicates batch normalization~\cite{Ioffe2015}. All leakyReLUs share the same ratio of 0.2 in the negative region. \lq\lq FC " is the fully connected layer.}
\label{tb:net}
\end{table}

\subsection{Comparison with state-of-the-art}
In this section, we compare our results with the state-of-the-art results. The results are shown in Table~\ref{tb:sota_mini} and~\ref{tb:sota_tiered}. 
For 1-shot tasks, the classification accuracy of our model achieves 1.52\% and 3.21\% improvements on miniImageNet and tieredImageNet respectively. 
For 5-shot tasks, the classification accuracy of our model achieves 1.11\% and 2.91\% improvements on miniImageNet and tieredImageNet respectively. 
Note that our model is quite simple and we perform no further adaptation to novel tasks. We use the same model for both 1-shot tasks and 5-shot tasks while most state-of-the-art models are trained separately for 1-shot tasks and 5-shot tasks. The results show that the generalization capability of our model is superior to previous top models.

\begin{table}[]
\begin{center}
\begin{tabular}{c|cc}
\hline
Model & 1-shot  & 5-shot  \\ \hline
SNAIL~\cite{mishra2017simple} & 55.71 $\pm$ 0.99\%  & 68.88 $\pm$ 0.92\% \\
Spyros~\etal~\cite{gidaris2018dynamic} & 56.20 $\pm$ 0.86\% & 73.00 $\pm$ 0.64\% \\
TADAM~\cite{oreshkin2018tadam} & 58.5 $\pm$ 0.3\% & 76.7 $\pm$ 0.3\% \\
TPN~\cite{liu2018learning} & 55.51 $\pm$ 0.86\% & 69.86$\pm$ 0.65\% \\
adaResNet~\cite{munkhdalai2018rapid} & 56.88 $\pm$ 0.62\% & 71.94 $\pm$ 0.57\% \\
Self-Jig~\cite{chen2019image} & 58.80 $\pm$ 1.36\% & 76.71 $\pm$ 0.72\% \\ 
CAML~\cite{jiang2018learning} & 59.23 $\pm$ 0.99\% & 72.35 $\pm$ 0.71\% \\
Qiao~\etal~\cite{qiao2018few} & 59.60 $\pm$ 0.41\% & 73.74 $\pm$ 0.19\% \\
LEO~\cite{rusu2018metalearning}   & 61.76 $\pm$ 0.08\%         & 77.59 $\pm$ 0.12\% \\
\hline
ours  & 63.28 $\pm$ 0.46 \%         & 78.70 $\pm$ 0.42\%         \\ 
\hline
\end{tabular}
\end{center}
\caption{Comparison with the state-of-the-art results with 95\% confidence intervals on miniImageNet 5-way classification.}
\label{tb:sota_mini}
\end{table}

\begin{table}[]
\begin{center}
\begin{tabular}{c|cc}
\hline
Model & 1-shot  & 5-shot  \\ \hline
MAML~\cite{liu2018learning} & 51.67 $\pm$ 1.81\%  & 70.30 $\pm$ 1.75\% \\
Prototypical Nets~\cite{ren2018meta} & 53.31 $\pm$ 0.89\% & 72.69 $\pm$ 0.74\% \\
Relation Net~\cite{liu2018learning} & 54.48 $\pm$ 0.93\% & 71.31 $\pm$ 0.78\% \\
TPN~\cite{liu2018learning} & 59.91 $\pm$ 0.94\% & 73.30 $\pm$ 0.75\% \\
LEO~\cite{rusu2018metalearning}   & 66.33 $\pm$ 0.05\%         & 81.44 $\pm$ 0.09 \% \\
\hline
ours  & 69.54 $\pm$ 0.52\%         & 84.35 $\pm$ 0.32\%         \\ 
\hline
\end{tabular}
\end{center}
\caption{Comparison with the state-of-the-art results with 95\% confidence intervals on tieredImageNet 5-way classification.}
\label{tb:sota_tiered}
\end{table}

\begin{figure*}
\centering
\includegraphics[width=1\linewidth]{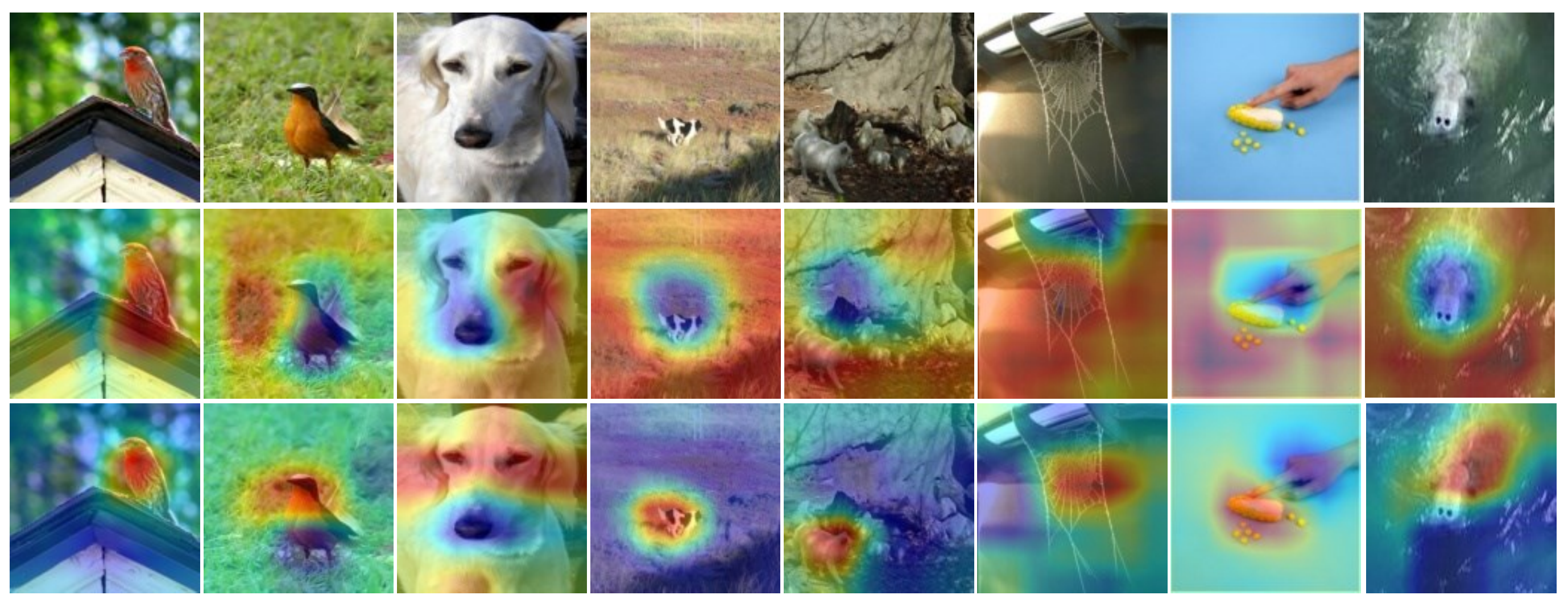} 
\caption{Adversarial region attention visualization. Red regions are positive regions that contribute to the classifier's correct prediction while the blue regions are adversarial regions that reduce classifier's confidence. The 1st row presents the input images. The 2nd row shows the adversarial region attention from C5-cls (see Section~\ref{sec:ab} for details) trained with low level features and the 3rd row shows the adversarial region attention from C5-C7-cls trained with multi-scale features. It can be seen from the 2nd row that model C5-cls incorrectly treats some background regions as the representative object. In the 3rd row, the object are more accurately attended to. (Best viewed in color.)
}
\label{fig:grad_57}
\end{figure*}

\subsection{Ablation study}
\label{sec:ab}
In order to show how much each component of our model contributes to the final performance. We conduct ablation studies in this section. Here we firstly introduce the baseline models that will be used for comparison. Although the training loss or the network architecture of baseline models are slightly different, layers from conv1 to conv5 which will be used in the test phase are the same as listed in Table~\ref{tb:net} for all models.

\begin{itemize}
\item \textbf{C5-cls}. The backbone CNN for this baseline model contains conv1 to conv5 layers. Feature maps from conv5 layer are global average pooled as feature vectors fed to the final linear classifier.  The training loss is the cross entropy loss and the model is trained without adversarial features. %
\item \textbf{C5-adv}. This model has the same backbone network as C5-cls. The difference is that this model is trained with the adversarial features. 
\item \textbf{C5-C7-cls}. This model has the same architecture with our model containing conv1 to conv7 layers. However, the model is trained using only multi-scale global average pooling features for classification. %

\end{itemize}

\begin{table}[]
\begin{center}
\begin{tabular}{l|ll|ll}
\hline
          & \multicolumn{2}{l|}{miniImageNet} & \multicolumn{2}{l}{tieredImageNet} \\ \hline
Model     & 1-shot & 5-shot & 1-shot & 5-shot           \\ 
C5-cls    & 58.24\% & 74.05\% & 66.21\% &                81.33\%  \\ 
C5-adv    & 61.56\%         & 76.97\%         &      68.90\%            &                 82.24\% \\ 
C5-C7-cls & 58.02\%         & 74.32\%         &   67.75\%               &       82.57\%           \\ \hline
ours      & 63.28\%         & 78.70\%          &     69.54\%             &       84.35\%           \\ \hline
\end{tabular}
\end{center}
\caption{Comparison with different baseline models on miniImageNet and tieredImageNet 5-way classification.}
\label{tb:as}
\end{table}

The comparison results are shown in Table~\ref{tb:as}. By comparing the classification accuracy of our model to that of C5-cls, we can find  large improvements on both 1-shot and 5-shot tasks demonstrating the effectiveness of adversarial feature learning and multi-scale feature learning. In the following subsections, we will evaluate the effectiveness of each component.

\subsubsection{Evaluation of adversarial attention}
By comparing C5-adv to C5-cls and our model to C5-C7-cls, we observe consistent improvements on both 1-shot and 5-shot tasks. The reason is that when trained with adversarial features, our model is able to explore the entire spatial area of the feature maps. Compared to focusing on conspicuous discriminative regions, spatial exploration provides chances to find more discriminative regions. The performance gap also provides evidence that (a) conventionally trained models tend to over-fit the training data and thus the generalization capability on novel classes is limited and (b) adversarial region attention can help reduce such over-fitting and increase the generalization capability without any change of the feature extraction part. 

\subsubsection{Evaluation of multi-scale feature learning}
For C5-cls and C5-C7-cls, there is a noticeable difference in their adversarial region attention. We illustrate the results in Figure~\ref{fig:grad_57}. From the figure, we can find that the classifier of C5-cls sometimes cannot capture the correct objects. It may mis-classify the background as the representative object. This is caused by the fact that the receptive field of the activations in conv5 is not large enough to capture the overall semantic discriminative features. When we use multi-scale feature learning, we observe that the attended regions are obviously more accurate. More accurate localization also helps adversarial feature learning which is justified by the fact that the classification accuracy of our model is higher than C5-adv on both datasets. 

\begin{figure}
\centering
\includegraphics[width=1\linewidth]{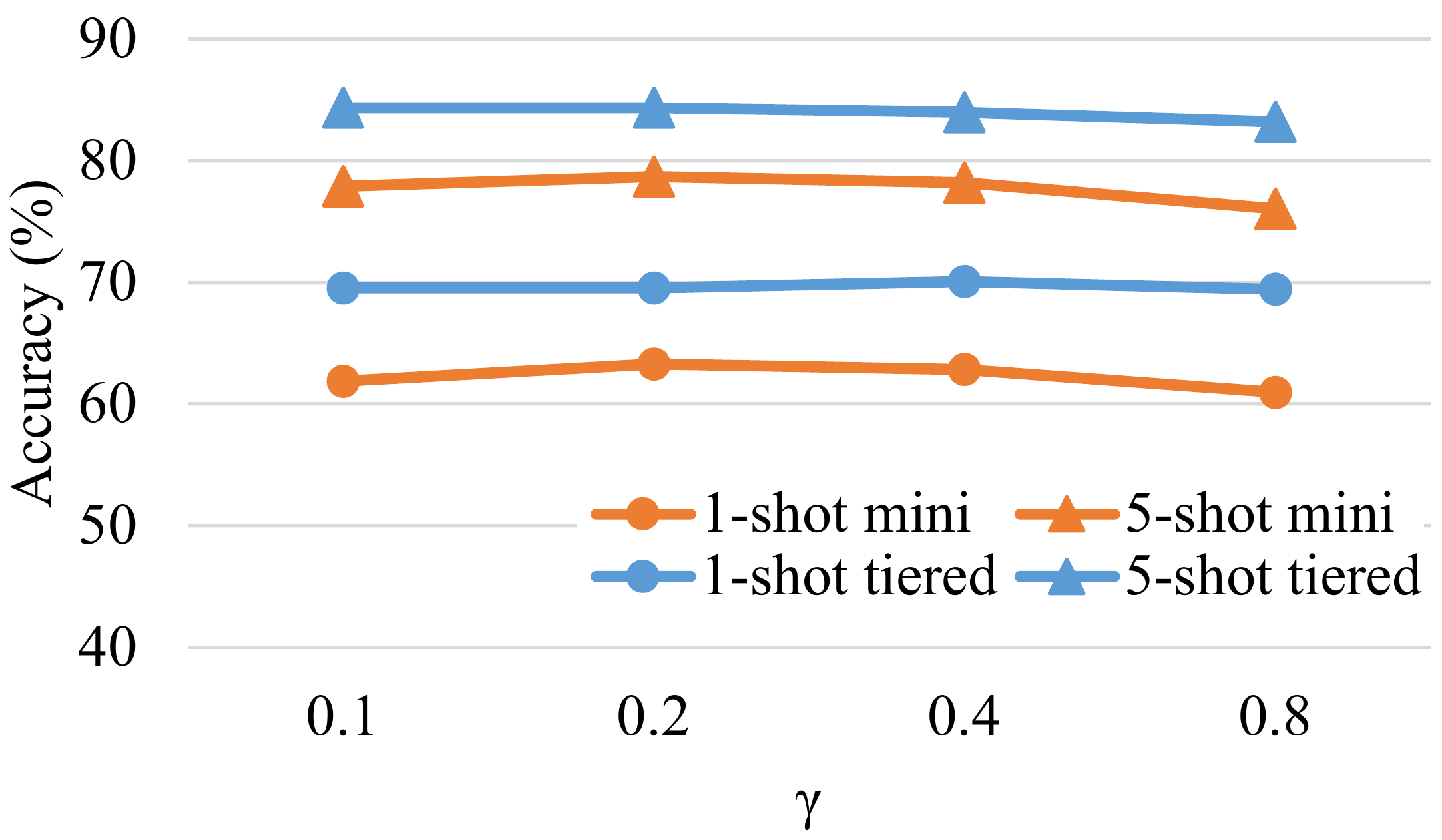} 
\caption{5-way 1/5-shot classification accuracy on miniImageNet and tieredImageNet with respect to different $\gamma$. 
}
\label{fig:diff_alpha}
\end{figure}

\subsubsection{Evaluation of varying step size $\gamma$}
\label{sec:gamma}
In Equation~\eqref{eq:adv_noise}, if $\gamma$ is too large, the averaged feature representation will be dominated and the large variance of feature vectors will cause the network too difficult to converge. However, a too small $\gamma$ will result in our model degrading to global average pooling. 
Therefore, we perform our evaluation of $\gamma$ in a reasonable range from 0.1 to 0.8 in Figure~\ref{fig:diff_alpha}. It can be found that within such a range the performance of our model is quite stable. The reason could be that the classifier is trained on multi-scale features in which the high level features are not perturbed. Thus they can serve as a reference for the alignment of adversarial features and thus contribute to the stable performance.

\begin{figure}
\centering
\includegraphics[width=1\linewidth]{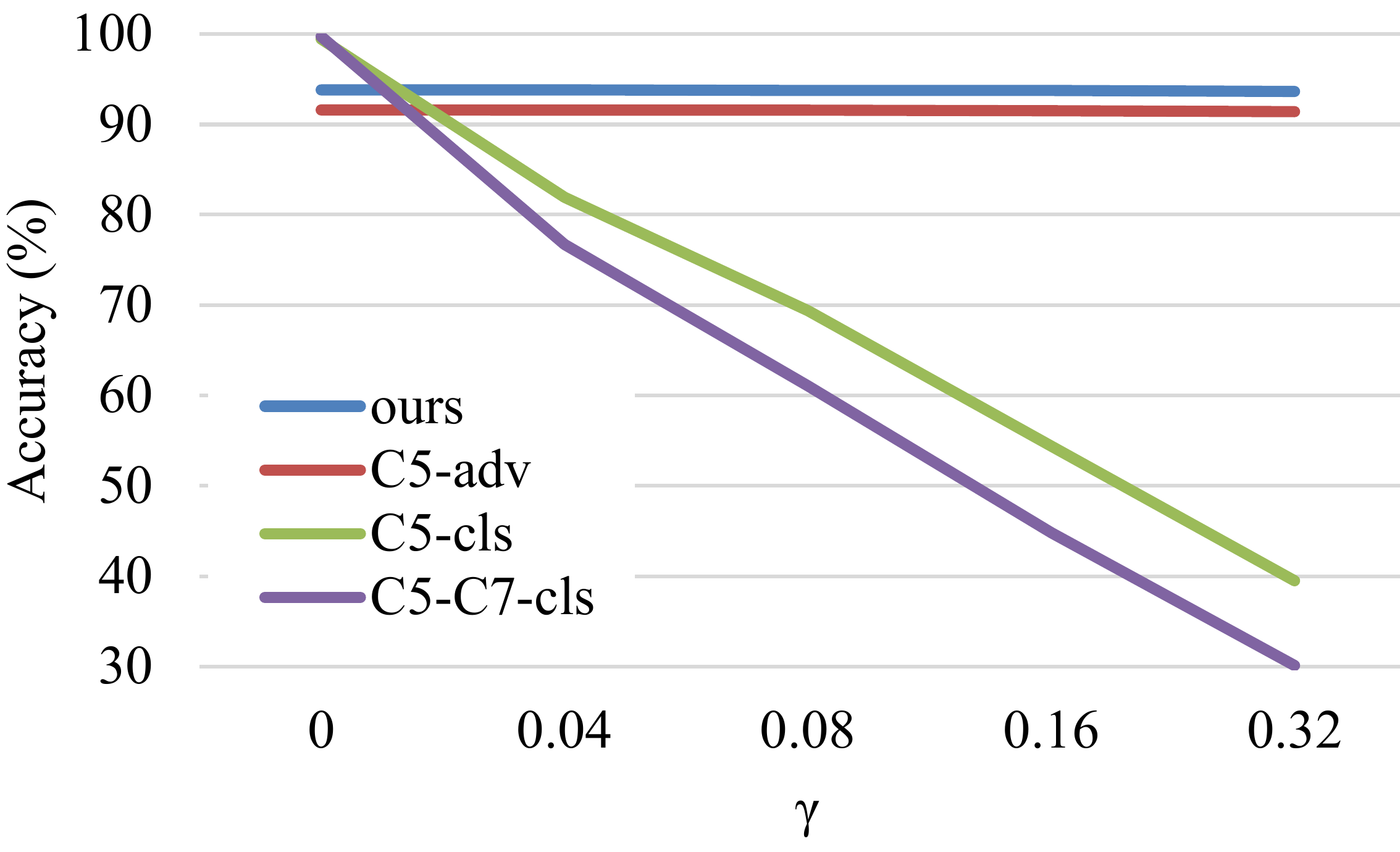} 
\caption{Classification accuracy on the training data of miniImageNet with respect to different levels of adversarial perturbations. The accuracies of models trained without adversarial features drop drastically even only a small amount of adversarial perturbations are added while models trained with adversarial features are much more robust to perturbations. (Best viewed in color.)
}
\label{fig:diff_adv}
\end{figure}

\subsubsection{Evaluation of model vulnerability}
In this section, we evaluate the vulnerability of baseline models when fed with adversarially perturbed features. Once we have trained all models, we gradually increase adversarial perturbations to construct adversarial features according to Equation~\eqref{eq:adv_noise} with increasing $\gamma$. The results are shown in Figure~\ref{fig:diff_adv}. The performance of C5-adv and our model are quite stable while the performance drops sharply for C5-cls and C5-C7-cls. It indicates that the conventionally trained model over-fits the training data such that a slight adversarial perturbation can cause incorrect prediction.

\subsubsection{Effect of dataset size}
In Table~\ref{tb:as}, the improvements on miniImageNet are 5.04\% and 4.65\% on 1-shot and 5-shot tasks while the improvements on tieredImageNet are 3.33\% and 3.02\% respectively. The difference in the improvements could be explained by the different size of the two datasets. tieredImageNet contains 351 classes for training while miniImageNet contains only 64 classes. With less diverse data, over-fitting problem can be more serious and thus impacts the model's ability to generalization. Therefore, the improvements on miniImageNet is larger than those on tieredImageNet. 

\section{Conclusion}
In this paper, we propose to learn generalizable features by learning from adversarial features. This approach is typically useful for tasks like few-shot classification where the test classes are different from training classes. 
Our model is quite simple and the trained feature extractor part can be easily combined with other methods. In our future work, we will try further adaptations like~\cite{gidaris2018dynamic} to see how much the performance can be improved. We will also try to automatically learn the scaling factor $s$ in cross entropy loss which is a fixed value in current work.

{\small
\bibliographystyle{ieee}
\bibliography{egbib}
}

\end{document}